%% file: 0_main.tex
\pdfoutput=1

\documentclass[11pt]{article}

\usepackage[final]{acl}

\usepackage{times}
\usepackage{latexsym}
\usepackage{amsmath}
\usepackage{amssymb}
\usepackage{booktabs}
\usepackage{tabularx}
\usepackage{graphicx}
\usepackage{xcolor,colortbl}
\usepackage{xspace}
\usepackage{mathtools, nccmath}
\usepackage{soul}
\usepackage{enumitem}
\usepackage[utf8]{inputenc}
\usepackage{csquotes}
\usepackage[capitalize]{cleveref}
\usepackage{anyfontsize}
\usepackage{multirow}

\usepackage[T1]{fontenc}
\usepackage[many]{tcolorbox}
\tcbuselibrary{listings}
\usetikzlibrary{arrows.meta}
\usetikzlibrary{calc}

\tcbuselibrary{skins}

\definecolor{pastelblue}{RGB}{173,216,230}
\definecolor{pastelyellow}{RGB}{255,253,208}
\definecolor{pastelpink}{RGB}{255,209,220}
\definecolor{pastelgreen}{RGB}{176,226,172}
\definecolor{pastellavender}{RGB}{230,230,250}
\definecolor{ForestGreen}{RGB}{34,139,34}
\newcommand\Mark[1]{\textsuperscript#1}

\usepackage[activate={true,nocompatibility}, kerning=true,spacing=true, stretch=10,shrink=10]{microtype}
\microtypecontext{spacing=nonfrench}

\makeatletter
\DeclareRobustCommand\onedot{\futurelet\@let@token\@onedot}
\def\@onedot{\ifx\@let@token.\else.\null\fi\xspace}

\def\eg{\textit{e.g}\onedot}

\def\vs{\textit{vs}\onedot}

\usepackage{inconsolata}

\title{\dataset: A Challenging Long-Context Mention Resolution Benchmark for LLMs}
\author{Kawshik Manikantan{\normalfont \Mark{1}}, Makarand Tapaswi{\normalfont \Mark{1}}, Vineet Gandhi{\normalfont \Mark{1}}, Shubham Toshniwal{\normalfont \Mark{2}}\\
\Mark{1}CVIT, IIIT Hyderabad \quad \Mark{2}NVIDIA \\
\small{ kawshik.manikantan@research.iiit.ac.in, \{makarand.tapaswi, vgandhi\}@iiit.ac.in, stoshniwal@nvidia.com }  
  }
 
\newcommand{\dataset}[0]{\texttt{IdentifyMe}\xspace}

\newcommand{\gptomini}[0]{\texttt{GPT-4o-mini}\xspace}
\newcommand{\gpto}[0]{\texttt{GPT-4o}\xspace}
\newcommand{\mistral}[0]{\texttt{Mistral-7B}\xspace}
\newcommand{\llama}[0]{\texttt{Llama-3.1-8B}\xspace}
\newcommand{\gemflash}[0]{\texttt{Gemini-1.5-Flash}\xspace}

\newcommand{\KM}[1]{{\color{blue} {\bf KM:} #1 }}
\newcommand{\MT}[1]{{\color{red} {\bf MT:} #1 }}
\newcommand{\ST}[1]{{\color{purple} {\bf ST:} #1 }}
\newcommand{\VG}[1]{{\color{orange} {\bf VG:} #1 }}

\renewcommand{\MT}[1]{}
\renewcommand{\ST}[1]{}
\renewcommand{\KM}[1]{}
\renewcommand{\VG}[1]{}

\begin{document}
\maketitle
\begin{abstract}

Recent evaluations of LLMs on coreference resolution have revealed that traditional output formats and evaluation metrics do not fully capture the models' referential understanding. 
To address this,  we introduce \dataset, a new benchmark for mention resolution presented in a multiple-choice question (MCQ) format, commonly used for evaluating LLMs.
\dataset features long narratives and employs heuristics to exclude easily identifiable mentions, creating a more challenging task.
The benchmark also consists of a curated mixture of different mention types and corresponding entities, allowing for a fine-grained model performance analysis. 
We evaluate both closed- and open-source LLMs on \dataset and observe a significant performance gap (20-30\%) between the state-of-the-art sub-10B open models \vs~closed ones.
We observe that pronominal mentions, which have limited surface information, are typically harder for models to resolve than nominal mentions. 
Additionally, we find that LLMs often confuse entities when their mentions overlap in nested structures.
The highest-scoring model, \gpto, achieves 81.9\% accuracy, highlighting the strong referential capabilities of state-of-the-art LLMs while also indicating room for further improvement.
\footnote{Code for the paper is available at:\\ \url{https://github.com/KawshikManikantan/IdentifyMe}}

\end{abstract}

\input{tex/1_intro}
\input{tex/2_dataset}

\input{tex/3_experiments}

\input{tex/4_results}

\input{tex/5_error_analysis}

\input{tex/6_conclusion}

\input{tex/7_ack}

\bibliography{custom}

\newpage
\appendix

\input{tex/appendix/0_main}

\end{document}

%% file: tex/1_intro.tex
\section{Introduction}
\input{figures/sample_instance}

Coreference Resolution (CR) consists of identifying the entity mentions and clustering them based on the entity identity.
It is a fundamental task for text comprehension and can therefore be used to assess a model's textual understanding.
While LLMs have made tremendous strides on a wide array of NLP tasks~\cite{brown2020language,openai2024gpt4,geminiteam2024}, their performance on CR has been relatively underwhelming.
It remains uncertain to what extent this is due to the LLMs’ weak referential abilities, as traditional coreference setups—both datasets and metrics—require LLMs to adhere to varying definitions of mentions, boundaries, and entities across datasets.

For instance, \citet{le2023large} report that on document-level coreference annotation, LLMs perform well at mention linking but struggle with mention detection, particularly due to varying definitions of what constitutes an entity and how mention boundaries are defined.
While \citet{manikantan-etal-2024-mei} mitigate the variability of entity definition by assuming major entities as inputs, their evaluation remains limited by dataset-specific span boundaries.
Recent work by \citet{gan-etal-2024-assessing} demonstrates through manual analysis that LLMs perform markedly better when evaluated in an unrestricted output mode.
This suggests that traditional evaluations may underestimate LLMs' coreference capabilities, highlighting the need to adapt traditional CR datasets and metrics to better assess LLMs. 

Along these lines, we introduce the \dataset benchmark for mention resolution in a multiple-choice question (MCQ) format. The MCQ format is commonly used in large language model (LLM) evaluations~\cite{hendrycks2021measuring} and offers two key advantages. First, its widespread presence in pretraining datasets enables LLMs to answer questions in this format effectively. Second, it eliminates the need for exact antecedent span identification during mention resolution evaluation, thus mitigating errors caused by dataset-specific annotation choices. 

To construct the benchmark, we use annotations from two long-text coreference benchmarks, namely LitBank~\cite{bamman-etal-2020-annotated} and FantasyCoref~\cite{han-etal-2021-fantasycoref}. 
To make the benchmark challenging, we restrict it to pronominal and nominal mentions and apply heuristics for each mention type to filter out easily resolvable cases (Section~\ref{sec:instance_selection}). 
Each MCQ instance consists of text marked with the mention of interest and choices comprising frequently occurring entities in the text and the \textit{None of the Above} (NoA) option.
~\cref{fig:sample_instance} shows an example in \dataset, derived from LitBank.

We evaluate both closed- and open-source LLMs with the following key findings:
\begin{itemize}\itemsep0em
    \item Among the mention types, LLMs perform worse on pronominal mentions (which have limited surface information) than on nominal mentions. 
    \item The instances where \emph{None of the Above} is the correct answer prove particularly challenging for all the models, with open-source models experiencing a performance drop of more than 50\%. 
    \item With nested mentions, LLMs frequently confuse entities with overlapping mentions (\eg,~\fbox{\fbox{his} mother}). 
    \item The highest-scoring model \gpto scores 81.9\% on \dataset, highlighting the strong performance of \emph{frontier} LLMs while indicating scope for further improvement in referential capabilities.  
\end{itemize}

%% file: figures/sample_instance.tex
\begin{figure}[t]
\centering
\includegraphics[width=0.95\linewidth]{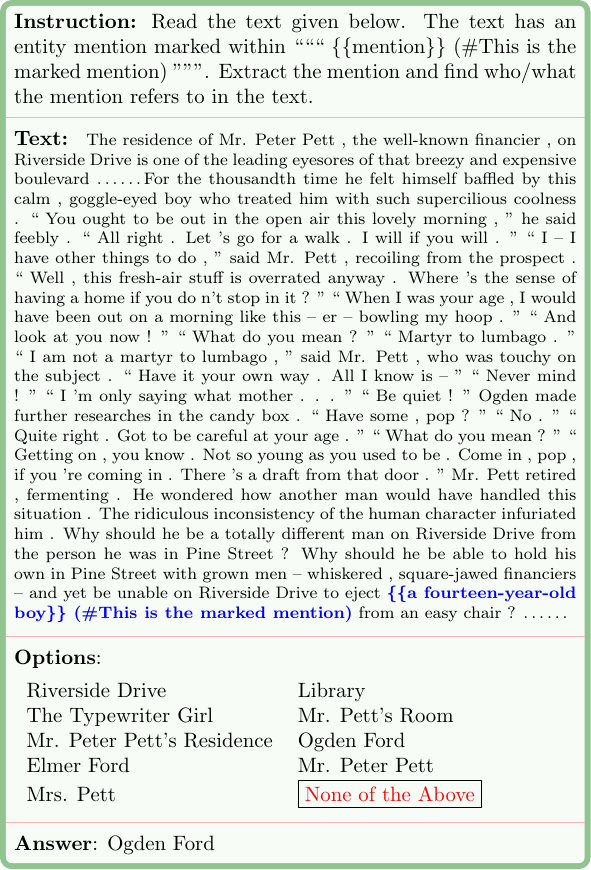}
\vspace{-1mm}
\caption{Sample instance from the validation set of \dataset. The mention of interest is highlighted in the text.
The answer options include frequently occurring entities in the text, and {None of the Above}.  
}
\vspace{-2mm}
\label{fig:sample_instance}
\end{figure}

%% file: tex/2_dataset.tex
\section{\dataset{} Benchmark}
\label{sec:dataset}

\dataset is an MCQ-based benchmark where, given a text document with a marked mention, the task is to identify the entity the mention refers to.
We derive these mentions from two coreference datasets focused on literary texts: LitBank and FantasyCoref.
These datasets provide long contexts (1700 words on average for FantasyCoref and 2000 words for LitBank) and feature multiple
entities with rich inter-dependencies (\eg,~\emph{Mr. and Mrs. Pett}) that make resolving mentions more challenging.
While LitBank offers diverse writing styles and linguistic structures,  FantasyCoref includes entities that often take on different forms  (\eg,~disguises and transformations),
or undergo title change (\eg, \textit{Prince Rudolph} is called \textit{The Emperor} after his coronation), which further complicates entity mapping.

Coreference annotations cluster mentions that refer to the same entity, but creating an MCQ requires a representative phrase for each entity cluster. 
We use \gptomini (see \cref{tab:prompt-label}) to generate these phrases based on the mentions and their frequencies.
The generated annotations undergo manual review to ensure each entity has a distinct representative phrase.

To prevent confusion, we discard and avoid labeling clusters that:
(i)~contain annotation errors (\eg, due to cluster merging or splitting~\cite{kummerfeld-klein-2013-error});
(ii)~are too small ($< 3$ mentions) or difficult or ambiguous to label (\eg,~entitites like \textit{some money});
(iii)~are plural, as they often lack explicit surface forms that can be derived from mentions.

An MCQ is created from a document using mentions from labeled clusters, with all labeled entities provided as options.
To ensure benchmark quality, we exclude short context documents ($< 1000$ words) or those where the discarded entities represent more than 50\% of the mentions.

\subsection{Selecting Mentions for \dataset}
\label{sec:instance_selection}

Based on previous works which utilize rule-based linguistic patterns to perform~\cite{zhou-su-2004-high, 10.1162/COLI_a_00152} or analyze~\cite{haghighi-klein-2009-simple,otmazgin-etal-2023-lingmess} coreference resolution, we propose a two-step heuristic to identify challenging mentions.

\paragraph{Step 1: Discard easy mentions.}
We apply two criteria to filter out mentions that can be easily resolved due to syntactic similarity:

{\emph{Nominal fuzzy score}}:
We calculate the fuzzy similarity\footnote{\url{https://github.com/seatgeek/thefuzz}} between a nominal mention and its entity's representative phrase, allowing for variations in word order and subsets.
We discard mentions with similarity scores above 75\%, as these cases typically provide obvious surface-form clues for identification.

{\emph{Net distractor score}}:
We categorize pronominal mentions based on attributes like gender, number, and animacy (LingMess~\cite{otmazgin-etal-2023-lingmess}).
For a candidate marked pronominal mention, nearby pronouns of the same category that refer to the same entity can provide disambiguating context.
However, pronouns that either share the category but refer to different entities, or refer to the same entity but have different categories, can increase ambiguity.
We define the \texttt{Net-Distractor-Score} as the difference between the count of ambiguity-increasing and disambiguating neighboring pronouns.
We discard mentions with non-positive scores ($\le 0$).

\paragraph{Step 2: Ranking mentions by difficulty.}
Filtered mentions are ranked from most to least difficult:
for nominals, a low \texttt{Nominal-Fuzzy-Score} is preferred; and
for pronouns, a high \texttt{Net-Distractor-Score} is preferred.
Additionally, the distance between the marked mention and other mentions of the same entity also indicate difficulty.
We consider distances to the nearest mention, the nearest nominal mention, and the nearest mention resembling the representative phrase as further criteria for ranking.
All the individual criteria are combined using Copeland's method~\cite{Copeland1951}, evaluating pairwise wins and losses to determine the final ranking.

\subsection{Dataset Statistics}

\input{tables/random}

\input{tables/human}
\input{tables/cot}
\input{tables/main}

\dataset comprises the 1800 most challenging questions based on our ranking method, drawn from 159 documents (64 from LitBank, 95 from FantasyCoref). 
We randomly select 600 of these questions as a validation set for prompt tuning and ablation experiments.
Each question includes a \textit{None of the Above (NoA)} option to encourage more confident entity selection.
To test NoA detection, we remove the correct entity from 10\% of the questions, making NoA the correct choice.
Both validation and test splits maintain balance across source datasets and mention types (pronominals and nominals).

\subsection{Does \dataset have Hard Mentions?}
We conduct an ablation experiment to assess the effectiveness of our mention selection process.
As a baseline, we randomly sample mentions and evaluate model performance on their identification.
The performance drops of 9.5\% for \mistral and 7.2\% for the more robust \gptomini demonstrate that \dataset captures more challenging mentions compared to random sampling (see \cref{tab:random}).

\subsection{Human Evaluation on \dataset{} Subset}

We perform human evaluation on a randomly selected subset of 10 FantasyCoref documents from the test split of \dataset.
A set of 50 mention resolution questions are extracted from these documents, comprising 25 nominals and 25 pronominal mentions.
As seen in~\cref{tab:human}, there is a significant performance gap of $\sim$23\% between humans and the best performing LLM, \gpto.
This confirms that there is substantial scope for improvement and \dataset{} poses a challenge to current LLMs.

\ST{We should just add the \gpto result? What do you think?}
\MT{while going through this text, found these messages, do we have time to fix? \\
Makarand: Shubham, what do you want to do about \gpto?
Shubham: We can skip it. I'm actually not completely sure about the random experiments. There are a few constraints which don't make sense IMO. Right now there are 10 subsets without any overlap. Ideally, it should've been 10 uniformly randomly chosen subsets. }

%% file: tables/random.tex
\begin{table}[t]
\centering
\small
\begin{tabular}{lcc}
\toprule
Model  & Random (10 runs)    & \dataset (Val.)   \\ \midrule
\mistral  & 64.8 $\pm$ 2.1 & 55.3 \\
\gptomini & 70.5 $\pm$ 1.9 & 63.3 \\
\gpto$^*$ & 83.8\phantom{ $\pm$ 1.9} & 80.7 \\
\bottomrule
\end{tabular}
\caption{Performance of models on the  \dataset validation set vs.\ comparable-sized evaluation set consisting of randomly chosen mentions (repeated 10 times). 
}
\label{tab:random}
\end{table}

%% file: tables/human.tex
\begin{table}[]
\centering
\small
\begin{tabular}{lc}
\toprule
Model/Approach & Accuracy \\ \midrule
\mistral & 46.0 \\
\llama & 50.0 \\ \midrule
\gptomini & 62.0 \\
\gemflash & 66.0 \\
\gpto & 70.0 \\ \midrule
Human-1 & 92.0 \\
Human-2 & 94.0 \\ \bottomrule
\end{tabular}
\caption{Performance of various models and human annotators on a subset of 50 questions from \dataset.
}
\label{tab:human}
\end{table}

%% file: tables/cot.tex
\begin{table}[t]
\centering
\small
\begin{tabular}{lcc}
\toprule
Model       & w/o CoT        & w/ CoT         \\ \midrule
\mistral  &   \textbf{55.3} & 53.8          \\
\llama & 50.2 &  \textbf{59.7}        \\ \midrule
\gptomini & 63.3          & \textbf{67.0} \\\bottomrule
\end{tabular}
\caption{Validation accuracy of LLMs w/ and w/o CoT.}
\label{tab:cot}
\end{table} 

%% file: tables/main.tex
\begin{table}[t]
\centering
\small
\tabcolsep=0.1cm
\begin{tabular}{lccc}
\toprule
Model &
  \begin{tabular}[c]{@{}c@{}}Total\\ (1200)\end{tabular} &
  \begin{tabular}[c]{@{}c@{}}Nominal\\ (600)\end{tabular} &
  \begin{tabular}[c]{@{}c@{}}Pronominal\\ (600)\end{tabular} \\ \midrule
Random           & \phantom{1}8.0           & \phantom{1}7.6             & \phantom{1}8.5            \\ \midrule
\mistral      & 51.5          & 52.5            & 50.5           \\
\llama      & 53.3          & 53.2            & 53.5           \\ \midrule
\gptomini      & 63.3          & 67.7            & 59.0           \\
\gemflash  & 73.9          & 77.7            & 70.0           \\
\gpto           & \textbf{81.9} & \textbf{85.2}   & \textbf{78.7}  \\\bottomrule
\end{tabular}
\caption{Performance of various models on the \dataset test set. }
\label{tab:main}
\end{table}

%% file: tex/3_experiments.tex
\section{Experiments}
\label{sec:experiments}

\paragraph{Models.} Among closed-source models, we evaluate
\gpto~\cite{openai2024gpt4},
\gptomini~\cite{openai2024gpt4omini}, and
\gemflash\footnote{Due to safety filters, evaluated on 1197 questions}~\cite{geminiteam2024}.
Due to computational constraints, we limit the evaluation of open-source models to sub-10B variants: \llama~\cite{dubey2024llama3herdmodels} and \mistral~\cite{jiang2023mistral}.

\paragraph{MCQ setup.}
The selected mention is highlighted in the original text by enclosing it with special tokens (\eg~``$\ldots$ eject \textit{a fourteen-year old boy} from $\ldots$''
$\rightarrow$
``$\ldots$ eject \{\{\textit{a fourteen-year old boy}\}\} (\#This is the marked span) from $\ldots$''.
A zero-shot prompt instructs the model to retrieve and resolve the mention and identify who or what it refers to from a given set of entities and NoA (detailed prompt in Appendix~\ref{sec:appendix_prompts}).

\paragraph{Inference details.}
For open-source models, we use regex-based constrained decoding with the \texttt{outlines} library~\cite{willard2023efficient} to limit answers to specific entity representative phrases.
We also experiment with a chain-of-thought (CoT) approach~\cite{wei2023chainofthought}, instructing the model to explain its reasoning before answering the question.
As seen in \cref{tab:cot}, using CoT improves the model performance (\eg, +9.5\% for \llama, +3.7\% for \gptomini). Based on these results, we use the CoT decoding for evaluation over the test set. For details on prompts used and decoding regular expressions, see Appendix~\ref{sec:appendix_prompts}.

%% file: tex/4_results.tex
\subsection{Results}
\cref{tab:main} presents the overall LLM performance on the \dataset test set, along with a breakdown by nominal and pronominal mention types.
The \texttt{Random} baseline, where answers are uniformly randomly chosen, achieves 8\% on our benchmark.
Although all LLMs outperform the \texttt{Random} baseline, open-source models show considerable room for improvement, with \llama reaching only 53.3\% accuracy.
\gpto is the top-performing model with an accuracy of 81.9\%.
Meanwhile, \gptomini, an affordable closed-source option, surpasses smaller open-source models but lags behind top performers like \gpto and \gemflash.
Across mention types, all closed-source models perform significantly better at resolving nominal mentions than pronominal ones.

\cref{tab:dataset} presents the performance split across mention types and source datasets. 
For nominal mentions, the FantasyCoref (FC) instances are, on average, considerably more challenging than those from LitBank (LB).
This could be because of the higher surface similarity across FantasyCoref entities (\eg~\textit{The eldest princess}, \textit{The youngest princess}).  
In contrast, LitBank's pronominal mentions are harder to resolve than FantasyCoref's, possibly due to its complex linguistic structure.

\input{tables/dataset}

\input{figures/error_nested_mention}

%% file: tables/dataset.tex
\begin{table}[]
\centering
\small
\begin{tabular}{@{}lcccc@{}}
\toprule
    & \multicolumn{2}{c}{Nominal} & \multicolumn{2}{c}{Pronominal}   \\ 
Model &
  \begin{tabular}[c]{@{}c@{}}FC\\ (300)\end{tabular} &
  \begin{tabular}[c]{@{}c@{}}LB\\ (300)\end{tabular} &
  \begin{tabular}[c]{@{}c@{}}FC\\ (300)\end{tabular} &
  \begin{tabular}[c]{@{}c@{}}LB\\ (300)\end{tabular} \\ \midrule
\mistral    & 39.0     & 66.0    & 51.7   & 49.3   \\
\llama      & 42.3     & 64.0    & 55.0   & 52.0   \\\midrule
\gptomini   & 60.7     & 74.7   & 63.3  & 54.7   \\
\gemflash   & 72.1     & 83.3   & 73.7  & 66.3   \\
\gpto   & \textbf{79.3} & \textbf{91.0} & \textbf{81.3} & \textbf{76.0} \\\bottomrule

\end{tabular}
\vspace{-2mm}
\caption{Performance split by mention type and dataset source. FC: FantasyCoref, LB: LitBank.
}
\label{tab:dataset}
\end{table}

%% file: figures/error_nested_mention.tex
\begin{figure}[t]
\centering
\includegraphics[width=0.8\linewidth]{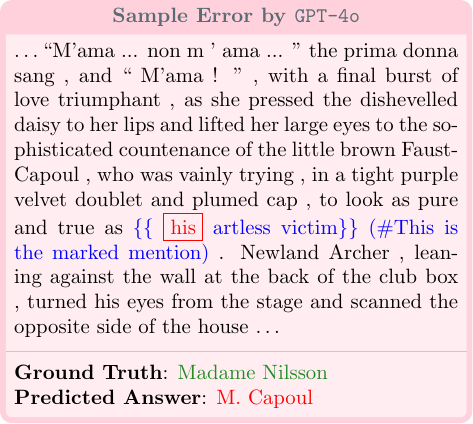}
\caption{An error by \gpto in resolving a nested mention where the model incorrectly resolves \textit{his artless victim} to the entity referred to by \textit{his} i.e.\ \textit{M. Capoul}.  
}
\vspace{-2mm}
\label{fig:sample_error_nested}
\end{figure}

%% file: tex/5_error_analysis.tex
\subsection{Error Analysis}

\input{tables/noa-comb}
\input{tables/nested}

\paragraph{Comparing entities \vs NoA.}
\cref{tab:noa-comb} provides the accuracy distribution when the correct option is an entity (Ent) \vs~NoA.
Furthermore, we classify errors into three categories:
(a)~ground truth is an entity and the model chooses another entity (Ent-Ent),
(b)~ground truth is an entity, but the model predicts NoA (Ent-NoA), and 
(c)~ground truth is NoA, but the model chooses an entity (NoA-Ent).
Open-source models perform extremely poorly on the NoA subset (120 MCQs), leading to high NoA-Ent errors.
Among closed-source models, \gemflash achieves sub-par performance on NoA MCQs ($\downarrow$ 48.3\%) and prefers to select an entity when the answer is NoA (83/120).
Interestingly, \gpto and \gptomini are much more resilient on NoA questions, with drops of only $\downarrow9.6\%$  and $\downarrow$ 0.9\%, respectively.

\paragraph{Nested mentions.}
The dataset contains 352 instances of nested mentions, where the span of one mention overlaps with another.
\cref{tab:nested} shows that the accuracy of nested mentions is comparable to the overall accuracy. 
However, when models err in resolving these mentions, about 40\% of these errors are because the predicted entity corresponds to an overlapping mention. 
Figure~\ref{fig:sample_error_nested} illustrates a sample nested mention error made by \gpto.

%% file: tables/noa-comb.tex
\begin{table}[t]
\centering
\small
\tabcolsep=0.1cm
\begin{tabular}{@{}lccccc@{}}
\toprule
                 & \multicolumn{2}{c}{Accuracy}  & \multicolumn{3}{c}{\#Misclassifications}            \\
Model            & Ent          & NoA  & Ent-Ent    & Ent-NoA   & NoA-Ent   \\
\midrule
\mistral       & 57.0          & \phantom{0}1.7           & 453          & \phantom{0}11          & 118         \\
\llama      & 59.2          & \phantom{0}0.8           & 438          & \phantom{00}3           & 119         \\ \midrule
\gptomini      & 63.4          & 62.5          & 221          & 174         & \phantom{0}45          \\
\gemflash & 78.6          & 30.3          & 192          & \phantom{0}38          & \phantom{0}83          \\
\gpto          & \textbf{82.9} & \textbf{73.3} & \textbf{135} & \phantom{0}\textbf{50} & \phantom{0}\textbf{32}
\\\bottomrule
\end{tabular}
\vspace{-1mm}
\caption{Left: Model accuracy for MCQs with correct answer as an entity (Ent, 1080 samples) \vs~NoA (120 samples).
Right: Number of misclassifications within entities (Ent-Ent) or with NoA (Ent-NoA, NoA-Ent).}
\label{tab:noa-comb}
\end{table}

%% file: tables/nested.tex
\begin{table}[]
\centering
\small

\begin{tabular}{lccc}

\toprule
\multirow{2}{*}{Model}           & \multicolumn{2}{c}{Accuracy} & \multirow{2}{*}{Span Error} \\
                &  Non-nested  & Nested    &         
\\\midrule

\mistral        & 50.1 & 54.8  & 40.3 \\
\llama          & 53.2 & 53.7  & 42.9 \\ \midrule
\gptomini       & 60.8 & 69.3  & 34.3 \\
\gemflash       & 73.3 & 75.1  & 36.8 \\
\gpto           & 82.1 & 81.5  & 47.7 \\\bottomrule
\end{tabular}
\vspace{-2mm}
\caption{LLM performance on nested mentions (352 of 1200) versus non-nested mentions. The Span Error column indicates the error for nested mentions where the predicted entity corresponds to an overlapping mention.   
}
\label{tab:nested}
\end{table}

%% file: tex/6_conclusion.tex
\section{Conclusion}
We present \dataset, a challenging MCQ benchmark designed for the evaluation of the referential capabilities of LLMs. Our analysis reveals several key challenges for LLMs, including: 
(i)~pronominal resolution which has limited surface form information, 
(ii)~questions where ``None of the Above" is the correct answer,
and
(iii)~nested mentions that require distinguishing between overlapping spans. 
\gpto scores 81.9\% on \dataset, highlighting the strong referential capabilities of \emph{frontier} LLMs while still leaving ample room for improvement. 
We believe the \dataset benchmark, with its curated mix of diverse and challenging mentions, will serve as an effective tool for fine-grained assessment of state-of-the-art LLMs' referential capabilities.

\section{Limitations}
The \dataset has several limitations: it covers only the literary domain, includes only nominal and pronominal mentions, and excludes plural entities. 
The source datasets we used are publicly available, and our preliminary investigations suggest limited contamination risk, as none of our evaluated LLMs could accurately reproduce the original CoNLL annotations for complete stories. 
While we significantly transformed the original coreference annotations to construct our benchmark, we acknowledge the potential possibility of data contamination.

%% file: tex/7_ack.tex
\section{Acknowledgements}
We are grateful to Varun Gupta for his assistance in setting up the annotation platform used for the user study. We also appreciate the valuable and timely contributions of the human annotators, Meenakshi and Sundar.

%% file: tex/appendix/0_main.tex
\appendix
\section{Appendix}
\label{sec:appendix}
\subsection{Model Details}
Table~\ref{tab:model-info} presents the precise model identifies used in this work.

\input{tables/model_identifier}

\input{figures/regex}
\subsection{Sample Model Outputs}

\begin{figure}[t]
\centering
\includegraphics[width=0.99\linewidth]{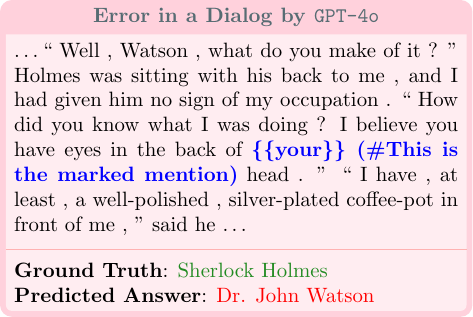}
\caption{A sample error made by \gpto where Sherlock Holmes and Dr.\ John Watson are engaged in a dialog. The instance is particularly hard because the dialog speakers are not marked and need to be inferred.
}
\label{fig:sample_error_dialog}
\end{figure}

Figure~\ref{fig:sample_error_dialog} presents another example error where the model is confused due to complicated first and second person references within dialog. 

\begin{figure}[t]
    \centering
    \includegraphics[width=\linewidth]{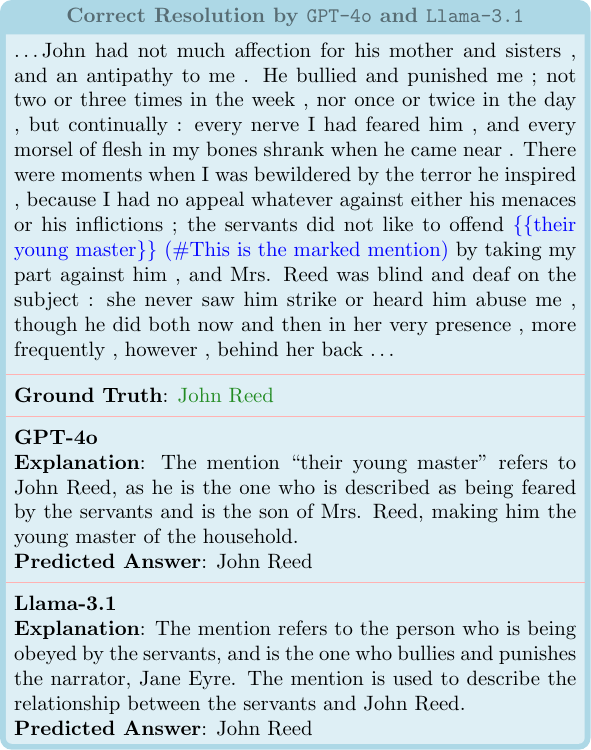}
    \caption{Sample instance from \dataset that both \texttt{GPT-4o} and \texttt{Llama-3.1} get right. }
    \label{fig:correct_instance_1}
\end{figure}

\begin{figure}[t]
    \centering
    \includegraphics[width=\linewidth]{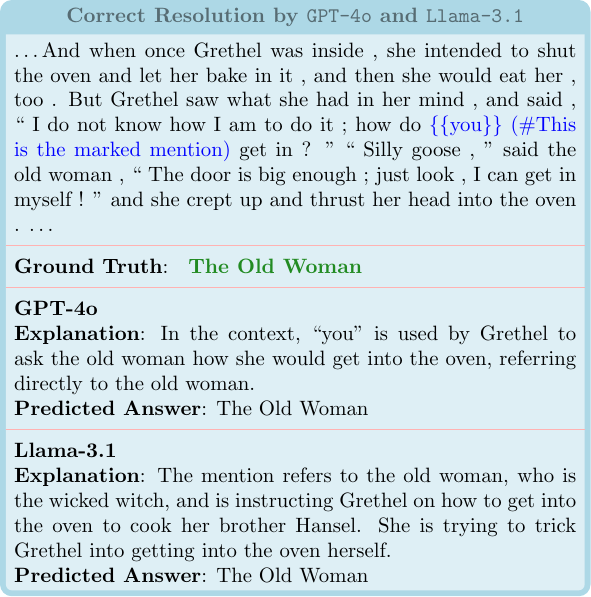}
    \caption{Sample instance from \dataset that both \texttt{GPT-4o} and \texttt{Llama-3.1} get right. }
    \label{fig:correct_instance_2}
\end{figure}

Figure~\ref{fig:correct_instance_1} and ~\ref{fig:correct_instance_2}  present instances which both \texttt{GPT-4o} and \texttt{Llama3.1} get right. Their explanation makes sense as well. 

\subsection{Prompts}
\label{sec:appendix_prompts}
Table~\ref{tab:prompt-label} presents the prompt passed to \gptomini to generate representative phrases for each cluster/entity. Table~\ref{tab:prompt-cot} and Table~\ref{tab:prompt-nocot} present the prompts and regular expressions used for inference with and without CoT respectively.
\input{tables/prompt-label}
\input{tables/prompt-cot}
\input{tables/prompt-nocot}

%% file: tables/model_identifier.tex
\begin{table}[h]
\centering
\small
\begin{tabular}{@{}ll@{}}
\toprule
\textbf{Model} & \textbf{Model Identifier} \\ \midrule
\mistral & \href{https://huggingface.co/mistralai/Mistral-7B-v0.3}{mistralai/Mistral-7B-v0.3} \\
\llama & \href{https://huggingface.co/meta-llama/Llama-3.1-8B}{meta-llama/Llama-3.1-8B} \\
\gptomini & \href{https://platform.openai.com/docs/models/gpt-4o-mini}{gpt-4o-mini-2024-07-18} \\
\gemflash & \href{https://cloud.google.com/vertex-ai/generative-ai/docs/learn/model-versions}{gemini-1.5-flash-001} \\
\gpto & \href{https://platform.openai.com/docs/models/gpt-4o}{gpt-4o-2024-08-06} \\ \bottomrule
\end{tabular}%
\vspace{-2mm}
\caption{Details of all the models evaluated in the paper.}
\label{tab:model-info}
\end{table}

%% file: tables/prompt-label.tex
\begin{table*}[t!]
\centering
\small
\ttfamily
\begin{tabular}{p{15cm}}
\toprule
\textbf{Instruction} \\
\midrule
You are provided with information about entities in a document. For each entity, you are given a list of different mentions, along with the number of occurrences of each mention in the format mention (count).
Derive an appropriate representative label for each entity from the given mentions.

Use the following guidelines:\\
- Prefer names over other noun phrases (nominals). \\
- If the entity appears to be a narrator but lacks a specific name, label the entity as "Narrator". \\
- Ensure the label is as precise and descriptive as possible. \\
- Avoid removing possessive pronouns from the representative label if they are included. \\
- Do not produce any other extra text. \\

Follow the below format: \\
Entity 0: Label 0 \\
Entity i: Label i \\ \midrule
\textbf{Example Input:} \\ \midrule
\textbf{Information:} \\
Entity 0: i(34), me(17), my(9), myself(3), ishmael(1), my soul(1) \\
Entity 1: the most absent-minded of men(1), that man(1) \\
Entity 2: an artist(1) \\
Entity 3: the commodore on the quarter-deck(1), their leaders(1) \\
Entity 4: your insular city of the manhattoes(1), the city of a dreamy sabbath afternoon(1) \\
Entity 5: the poor poet of tennessee(1) \\
Entity 6: the world(2), this world(1) \\
Entity 7: cato(1) \\
Entity 8: this shepherd(1), the shepherd(1) \\
Entity 9: narcissus(1) \\ \midrule
\textbf{Example Output:} \\ \midrule
Entity 0: Ishmael  \\
Entity 1: The Most Absent-Minded Man \\  
Entity 2: An Artist  \\
Entity 3: The Commodore  \\
Entity 4: City of the Manhattoes  \\
Entity 5: The Poor Poet of Tennessee  \\
Entity 6: The World  \\
Entity 7: Cato  \\
Entity 8: The Shepherd  \\
Entity 9: Narcissus \\ \midrule
\end{tabular}%
\caption{The zero-shot prompt passed to \gptomini to generate representative phrases for each cluster/entity.}
\label{tab:prompt-label}
\end{table*}

%% file: tables/prompt-cot.tex
\begin{table*}[t!]
\centering
\small
\ttfamily
\begin{tabular}{p{15cm}}
\toprule
\textbf{Instruction} \\
\midrule
Read the text given below. The text has an entity mention marked within """ \{\{mention\}\} (\#This is the marked mention) """. Extract the mention and find who/what the mention refers to in the text.\\ \midrule
\textbf{Example Input:} \\ \midrule
\textbf{Text:} \\
Chapter 1 It is a truth universally acknowledged, that a single man in possession of a good fortune, must be in want of a wife. However little known the feelings or views of such a man may be on his first entering a neighbourhood, this truth is so well fixed in the minds of the surrounding families, that he is considered the rightful property of some one or other of their daughters. "My dear Mr. Bennet," said his lady to him one day, \ldots \par
\\
Chapter 2 Mr. Bennet was among the earliest of those who waited on Mr. Bingley. He had always intended to visit him, though to the last always assuring {\color{red} \{\{his wife\}\} (\#This is the marked mention)} that he should not go; and till the evening after the visit was paid she had no knowledge of it. It was then disclosed in the following manner. Observing his second daughter employed in trimming a hat, he suddenly addressed her with: "I hope Mr. Bingley will like it, Lizzy." "We are not in a way to know what Mr. Bingley likes," said her mother resentfully, "since we are not to visit" \ldots \par
\\
I do not know how you will ever make him amends for his kindness; or me, either, for that matter. At our time of life it is not so pleasant, I can tell you, to be making new acquaintances every day; but for your sakes, we would do anything. Lydia, my love, though you are the youngest, I dare say Mr. Bingley will dance with you at the next ball. \par
\\

\textbf{Options for the answer:} \\ \par
Mary \\
Kitty \\
Mrs. Bennet \\
Mrs. Long \\
Elizabeth \\
Mr. Bingley \\
Mr. Bennet \\
Lydia \\
Netherfield Park \\
None of the Above \\ \par
\\

Note that you can select the None of the Above option (The mention refers to: None of the Above), if the mention does not refer to any other entity/option. Also provide explanations in 1-2 sentences for the same. Do not produce any other extra text.\\
Follow the below format:\\
- Mention: \\
- Explanation: \\
- The mention refers to: \\ \par
\\ \midrule
\textbf{Decoding Regex (Constrained Decoding):} \\ \midrule
\texttt{- Mention: \textbackslash\{\{[A-Za-z ,\textbackslash'\textbackslash.]\{1,125\}\textbackslash\}\}\newline- Explanation: \textbackslash\{\{[A-Za-z ,\textbackslash'\textbackslash.]\{150,350\}\textbackslash\}\}\newline- The mention refers to: (Mary|Kitty|Mrs. Bennet|Mrs. Long|Elizabeth|Mr. Bingley|\newline Mr. Bennet|Lydia|Netherfield Park|None of the Above)}\\ \midrule
\textbf{Example Output:} \\ \midrule
- Mention: {{his wife}} \\
- Explanation: The mention refers to Mrs. Bennet. The pronoun 'his' refers to Mr. Bennet, and 'wife' refers to the person who is married to Mr. Bennet. So, the mention refers to Mrs. Bennet.\\
- The mention refers to: Mrs. Bennet \\\bottomrule
\end{tabular}%
\caption{QA prompt with CoT used in the test set experiments.}
\label{tab:prompt-cot}
\end{table*}

%% file: tables/prompt-nocot.tex
\begin{table*}[t!]
\centering
\small
\ttfamily
\begin{tabular}{p{15cm}}
\toprule
\textbf{Instruction} \\
\midrule
Read the text given below. The text has an entity mention marked within """ \{\{mention\}\} (\#This is the marked mention) """. Extract the mention and find who/what the mention refers to in the text.\\ \midrule
\textbf{Example Input:} \\ \midrule
\textbf{Text:} \\
Chapter 1 It is a truth universally acknowledged, that a single man in possession of a good fortune, must be in want of a wife. However little known the feelings or views of such a man may be on his first entering a neighbourhood, this truth is so well fixed in the minds of the surrounding families, that he is considered the rightful property of some one or other of their daughters. "My dear Mr. Bennet," said his lady to him one day, \ldots \par
\\
Chapter 2 Mr. Bennet was among the earliest of those who waited on Mr. Bingley. He had always intended to visit him, though to the last always assuring {\color{red} \{\{his wife\}\} (\#This is the marked mention)} that he should not go; and till the evening after the visit was paid she had no knowledge of it. It was then disclosed in the following manner. Observing his second daughter employed in trimming a hat, he suddenly addressed her with: "I hope Mr. Bingley will like it, Lizzy." "We are not in a way to know what Mr. Bingley likes," said her mother resentfully, "since we are not to visit" \ldots \par
\\
I do not know how you will ever make him amends for his kindness; or me, either, for that matter. At our time of life it is not so pleasant, I can tell you, to be making new acquaintances every day; but for your sakes, we would do anything. Lydia, my love, though you are the youngest, I dare say Mr. Bingley will dance with you at the next ball. \par
\\

\textbf{Options for the answer:} \\ \par
Mary \\
Kitty \\
Mrs. Bennet \\
Mrs. Long \\
Elizabeth \\
Mr. Bingley \\
Mr. Bennet \\
Lydia \\
Netherfield Park \\
None of the Above \\ \par
\\

Note that you can select the None of the Above option (The mention refers to: None of the Above), if the mention does not refer to any other entity/option. Do not produce any other extra text.\\
Follow the below format:\\
- Mention: \\
- The mention refers to: \\ \par
\\ \midrule
\textbf{Decoding Regex (Constrained Decoding):} \\ \midrule
\texttt{- Mention: \textbackslash\{\{[A-Za-z ,\textbackslash'\textbackslash.]\{1,125\}\textbackslash\}\}\newline- The mention refers to: (Mary|Kitty|Mrs. Bennet|Mrs. Long|Elizabeth|Mr. Bingley|\newline Mr. Bennet|Lydia|Netherfield Park|None of the Above)}\\ \midrule
\textbf{Example Output:} \\ \midrule
- Mention: {{his wife}} \\
- The mention refers to: Mrs. Bennet \\\bottomrule
\end{tabular}%
\caption{QA prompt without CoT.}
\label{tab:prompt-nocot}
\end{table*}

%% file: 0_main.bbl
\begin{thebibliography}{20}
\expandafter\ifx\csname natexlab\endcsname\relax\def\natexlab#1{#1}\fi

\bibitem[{Bamman et~al.(2020)Bamman, Lewke, and Mansoor}]{bamman-etal-2020-annotated}
David Bamman, Olivia Lewke, and Anya Mansoor. 2020.
\newblock {An Annotated Dataset of Coreference in {E}nglish Literature}.
\newblock In \emph{LREC}.

\bibitem[{Brown et~al.(2020)Brown, Mann, Ryder, Subbiah, Kaplan, Dhariwal, Neelakantan, Shyam, Sastry, Askell, Agarwal, Herbert-Voss, Krueger, Henighan, Child, Ramesh, Ziegler, Wu, Winter, Hesse, Chen, Sigler, Litwin, Gray, Chess, Clark, Berner, McCandlish, Radford, Sutskever, and Amodei}]{brown2020language}
Tom~B. Brown, Benjamin Mann, Nick Ryder, Melanie Subbiah, Jared Kaplan, Prafulla Dhariwal, Arvind Neelakantan, Pranav Shyam, Girish Sastry, Amanda Askell, Sandhini Agarwal, Ariel Herbert-Voss, Gretchen Krueger, Tom Henighan, Rewon Child, Aditya Ramesh, Daniel~M. Ziegler, Jeffrey Wu, Clemens Winter, Christopher Hesse, Mark Chen, Eric Sigler, Mateusz Litwin, Scott Gray, Benjamin Chess, Jack Clark, Christopher Berner, Sam McCandlish, Alec Radford, Ilya Sutskever, and Dario Amodei. 2020.
\newblock {Language Models are Few-Shot Learners}.
\newblock In \emph{NeurIPS}.

\bibitem[{Copeland(1951)}]{Copeland1951}
A.~Copeland. 1951.
\newblock {A Reasonable Social Welfare Function}.
\newblock In \emph{Seminar on Applications of Mathematics to Social Sciences}.

\bibitem[{Gan et~al.(2024)Gan, Poesio, and Yu}]{gan-etal-2024-assessing}
Yujian Gan, Massimo Poesio, and Juntao Yu. 2024.
\newblock {Assessing the Capabilities of Large Language Models in Coreference: An Evaluation}.
\newblock In \emph{LREC-COLING}.

\bibitem[{{Gemini Team} et~al.(2024){Gemini Team}, Georgiev, and Lei}]{geminiteam2024}
{Gemini Team}, Petko Georgiev, and Ving~Ian Lei. 2024.
\newblock \href {http://arxiv.org/abs/2403.05530} {{Gemini 1.5: Unlocking multimodal understanding across millions of tokens of context}}.

\bibitem[{Haghighi and Klein(2009)}]{haghighi-klein-2009-simple}
Aria Haghighi and Dan Klein. 2009.
\newblock {Simple Coreference Resolution with Rich Syntactic and Semantic Features}.
\newblock In \emph{EMNLP}.

\bibitem[{Han et~al.(2021)Han, Seo, Kang, Kim, Choi, Song, and Choi}]{han-etal-2021-fantasycoref}
Sooyoun Han, Sumin Seo, Minji Kang, Jongin Kim, Nayoung Choi, Min Song, and Jinho~D. Choi. 2021.
\newblock {{F}antasy{C}oref: Coreference Resolution on Fantasy Literature Through Omniscient Writer{'}s Point of View}.
\newblock In \emph{Fourth Workshop on Computational Models of Reference, Anaphora and Coreference}.

\bibitem[{Hendrycks et~al.(2021)Hendrycks, Burns, Basart, Zou, Mazeika, Song, and Steinhardt}]{hendrycks2021measuring}
Dan Hendrycks, Collin Burns, Steven Basart, Andy Zou, Mantas Mazeika, Dawn Song, and Jacob Steinhardt. 2021.
\newblock \href {https://openreview.net/forum?id=d7KBjmI3GmQ} {{Measuring Massive Multitask Language Understanding}}.
\newblock In \emph{ICLR}.

\bibitem[{Jiang et~al.(2023)Jiang, Sablayrolles, Mensch, Bamford, Chaplot, Casas, Bressand, Lengyel, Lample, Saulnier et~al.}]{jiang2023mistral}
Albert~Q Jiang, Alexandre Sablayrolles, Arthur Mensch, Chris Bamford, Devendra~Singh Chaplot, Diego de~las Casas, Florian Bressand, Gianna Lengyel, Guillaume Lample, Lucile Saulnier, et~al. 2023.
\newblock {Mistral 7B}.
\newblock \emph{arXiv preprint arXiv:2310.06825}.

\bibitem[{Kummerfeld and Klein(2013)}]{kummerfeld-klein-2013-error}
Jonathan~K. Kummerfeld and Dan Klein. 2013.
\newblock {Error-Driven Analysis of Challenges in Coreference Resolution}.
\newblock In \emph{EMNLP}.

\bibitem[{Le and Ritter(2023)}]{le2023large}
Nghia~T. Le and Alan Ritter. 2023.
\newblock {Are Large Language Models Robust Coreference Resolvers?}
\newblock \emph{arXiv preprint arXiv:22305.14489}.

\bibitem[{Lee et~al.(2013)Lee, Chang, Peirsman, Chambers, Surdeanu, and Jurafsky}]{10.1162/COLI_a_00152}
Heeyoung Lee, Angel Chang, Yves Peirsman, Nathanael Chambers, Mihai Surdeanu, and Dan Jurafsky. 2013.
\newblock \href {https://doi.org/10.1162/COLI_a_00152} {{Deterministic Coreference Resolution Based on Entity-Centric, Precision-Ranked Rules}}.
\newblock \emph{Computational Linguistics}.

\bibitem[{Manikantan et~al.(2024)Manikantan, Toshniwal, Tapaswi, and Gandhi}]{manikantan-etal-2024-mei}
Kawshik Manikantan, Shubham Toshniwal, Makarand Tapaswi, and Vineet Gandhi. 2024.
\newblock {Major Entity Identification: A Generalizable Alternative to Coreference Resolution}.
\newblock In \emph{EMNLP}.

\bibitem[{Meta-AI(2024)}]{dubey2024llama3herdmodels}
Meta-AI. 2024.
\newblock \href {http://arxiv.org/abs/2407.21783} {{The Llama 3 Herd of Models}}.

\bibitem[{OpenAI(2024{\natexlab{a}})}]{openai2024gpt4}
OpenAI. 2024{\natexlab{a}}.
\newblock \href {http://arxiv.org/abs/2303.08774} {{GPT-4 Technical Report}}.

\bibitem[{OpenAI(2024{\natexlab{b}})}]{openai2024gpt4omini}
OpenAI. 2024{\natexlab{b}}.
\newblock \href {https://openai.com/index/gpt-4o-mini-advancing-cost-efficient-intelligence/} {{GPT-4o-mini: Advancing Cost-Efficient Intelligence}}.

\bibitem[{Otmazgin et~al.(2023)Otmazgin, Cattan, and Goldberg}]{otmazgin-etal-2023-lingmess}
Shon Otmazgin, Arie Cattan, and Yoav Goldberg. 2023.
\newblock {{L}ing{M}ess: Linguistically Informed Multi Expert Scorers for Coreference Resolution}.
\newblock In \emph{EACL}.

\bibitem[{Wei et~al.(2023)Wei, Wang, Schuurmans, Bosma, Ichter, Xia, Chi, Le, and Zhou}]{wei2023chainofthought}
Jason Wei, Xuezhi Wang, Dale Schuurmans, Maarten Bosma, Brian Ichter, Fei Xia, Ed~Chi, Quoc Le, and Denny Zhou. 2023.
\newblock \href {https://arxiv.org/abs/2201.11903} {{Chain-of-Thought Prompting Elicits Reasoning in Large Language Models}}.
\newblock In \emph{NeurIPS}.

\bibitem[{Willard and Louf(2023)}]{willard2023efficient}
Brandon~T Willard and R{\'e}mi Louf. 2023.
\newblock {Efficient Guided Generation for LLMs}.
\newblock \emph{arXiv preprint arXiv:2307.09702}.

\bibitem[{Zhou and Su(2004)}]{zhou-su-2004-high}
GuoDong Zhou and Jian Su. 2004.
\newblock {A High-Performance Coreference Resolution System using a Constraint-based Multi-Agent Strategy}.
\newblock In \emph{COLING}.

\end{thebibliography}
